\documentclass[runningheads]{llncs}
\usepackage[T1]{fontenc}
\usepackage{multirow}
\usepackage{multicol}

\usepackage{cite}
\usepackage{amsmath,amssymb,amsfonts}
\usepackage{algorithmic}
\usepackage{graphicx}
\usepackage{textcomp}
\usepackage{makecell}
\usepackage[normalem]{ulem}
\usepackage{xcolor}
\usepackage{booktabs}
\usepackage[colorlinks=true, allcolors=green]{hyperref}
\usepackage{subfigure}
\usepackage{colortbl} % 用于表格着色
\usepackage{xcolor} % 定义颜色
\begin{document}
\title{MBDS: A Multi-Body Dynamics Simulation Dataset for Graph Networks Simulators}
\titlerunning{MBDS: A Multi-Body Dynamics Simulation Dataset}
\author{Sheng Yang\inst{1,2}  \and
Fengge Wu\inst{2}\and Junsuo Zhao\inst{2}}
\institute{University of Chinese Academy of Sciences \and
Institute of Software, Chinese Academy of Sciences\\
\email{\{yangsheng2023, jiadaixi2022, fenggewu, junsuozhao\}@iscas.ac.cn}}
\maketitle              % typeset the header of the contribution
\begin{abstract}
Modeling the structure and events of the physical world constitutes a fundamental objective of neural networks. Among the diverse approaches, Graph Network Simulators (GNS) have emerged as the leading method for modeling physical phenomena, owing to their low computational cost and high accuracy. The datasets employed for training and evaluating physical simulation techniques are typically generated by researchers themselves, often resulting in limited data volume and quality. Consequently, this poses challenges in accurately assessing the performance of these methods.
In response to this, we have constructed a high-quality physical simulation dataset encompassing 1D, 2D, and 3D scenes, along with more trajectories and time-steps compared to existing datasets. Furthermore, our work distinguishes itself by developing eight complete scenes, significantly enhancing the dataset’s comprehensiveness. A key feature of our dataset is the inclusion of precise multi-body dynamics, facilitating a more realistic simulation of the physical world. Utilizing our high-quality dataset, we conducted a systematic evaluation of various existing GNS methods. Our dataset is accessible for download at https://github.com/Sherlocktein/MBDS, offering a valuable resource for researchers to enhance the training and evaluation of their methodologies.

%Graph Neural Networks (GNNs) have emerged as a popular tool for learning physical dynamics, requiring only a fraction of the computational cost of traditional simulators. While current methods have achieved notable success in tackling a broad spectrum of physical problems, a major challenge that persists in this domain is the limited availability of comprehensive datasets. These datasets are crucial for establishing unified benchmarks, especially in the context of multi-body dynamics scenarios, which are frequently encountered in real-world applications. To fill an existing timely gap and foster further advancements in research, we propose a Multi-Body Dynamics Simulation dataset (MBDS) designed to more accurately simulate real-life scenarios. The MBDS dataset encompasses numerous diverse yet common scenarios. Each scenario consists of 15000 trajectories with 200 time-steps. Furthermore, it easily enables the generation of your desired scenarios based on the foundational work we have laid out. We perform extensive experiments using our dataset to assess the predictive accuracy of various Graph Network Simulators (GNS) models, aiding in the analysis of issues in these models. We anticipate that the release of the MBDS dataset will become a benchmark for comparing various models, and will further promote research and development in the field of GNS.
\end{abstract}


\section{Introduction}
% Modeling and simulation of physical systems play a crucial role in advancing scientific understanding, technological innovation, and problem-solving across various domains \cite{}. Several differentiable simulators based on neural networks have been proposed \cite{}, and they have demonstrated notable success in the accurate simulation of diverse types of interacting entities. The advent of these methodologies has facilitated the completion of physics simulations through end-to-end trained neural networks. Unlike neural network models utilized in other domains such as video\cite{}, image\cite{}, and text\cite{}, the introduction of these methodologies has pioneered the potential to model and analyze the physical world based on neural networks, thereby opening up new avenues for the application and expansion of deep learning methods. 

Modeling and simulating physical systems play an important role in advancing scientific comprehension, technological innovation, and cross-disciplinary problem-solving \cite{physics}. Numerous differentiable simulators, leveraging neural network frameworks, have been proposed in the literature \cite{HGNN, LGNN, SGNN, GNS }, showcasing considerable efficacy in the precise emulation of a wide spectrum of interacting entities. The inception of these methodologies has not only streamlined the execution of physics simulations but has also harnessed the capabilities of end-to-end trained neural networks for heightened accuracy. In contrast to neural network models applied in alternative domains such as video \cite{Physion}, image \cite{image}, and text \cite{text}, the incorporation of these methodologies represents a pioneering initiative toward modeling and analyzing the physical world utilizing neural networks. This paradigm shift not only augments the depth of simulation fidelity but also unfolds novel prospects for the application and extension of deep learning methods in the realm of physical system modeling and simulation.

\begin{table*}[t]
\renewcommand\arraystretch{1.1}
\caption{Comparison across related datasets. ``--'' means is not officially reported in the corresponding paper.}
\label{tab:comparision}
\definecolor{mygray}{rgb}{0.8, 0.8, 0.8} % 浅灰色

\centering
\scalebox{0.78}{
\begin{tabular}{@{}lccc|cccc|cc|cc@{}}
\toprule
\hline
\textbf{Dataset} & \multicolumn{3}{c}{\textbf{Dimensions}}  & \multicolumn{4}{c}{ \textbf{The Number of}} & \multicolumn{2}{c}{\textbf{Force}}& \multicolumn{2}{c}{\textbf{Dynamics}} \\

 &1D &2D&3D  & Trajectories & Time-steps  & Scenarios & Objects & Static  & Variable & Single&Multi  \\
 \hline
\bottomrule
RigidFall\cite{RigidFall} & \texttimes & \texttimes & \checkmark & 5000 & 120& 1 & 3 & \checkmark & \texttimes  &\checkmark &\texttimes \\
FLAGSIMPLE \cite{GNS} & \texttimes & \checkmark & \texttimes  & 800 & 200& 1 & 1 & \texttimes & \texttimes  & \checkmark &\texttimes  \\
CYLINDERFLOW \cite{GNS}  &\texttimes & \checkmark & \texttimes & 800 & 400& 1 & 1 & \texttimes & \texttimes & \checkmark &\texttimes \\
AIRFOIL\cite{Mesh/iclr/PfaffFSB21}     &\texttimes & \checkmark & \texttimes & 1000 & 150 & 1 & 1 &\texttimes & \texttimes & \checkmark &\texttimes \\
Cub-Toss \cite{Cubtoss}    &\texttimes & \texttimes & \checkmark  & 10000 & 80 & 1 & 1 &\texttimes & \texttimes & \checkmark &\texttimes \\
INFLATINGFONT \cite{BMS}   &\texttimes & \checkmark & \texttimes  & 1000 & 100 & 1 & 1 &\texttimes & \texttimes  & \checkmark &\texttimes  \\
\hline
N-Pendulum \cite{LGNN} &\texttimes & \checkmark & \texttimes & 100 & 100 & 5 &1 &\checkmark & \texttimes  &\checkmark &\texttimes\\
N-Spring \cite{HGNN}  &\texttimes & \checkmark & \texttimes & 100 & 100 & 5 & 1 &\checkmark & \texttimes  &\checkmark &\texttimes\\
Spring-ball \cite{LGNN} &\texttimes & \checkmark & \texttimes & 1000 & 50 & 1 & 2 &\texttimes &\checkmark  &\texttimes &\checkmark\\

MD17 \cite{MD17}   &\texttimes & \checkmark & \texttimes & 2000 & 500 & 1 & 2  & \texttimes &\checkmark  &\texttimes &\checkmark\\
CMU\cite{CMU} &\texttimes & \texttimes & \checkmark  & 86 & 100 & 1 & 1 &\texttimes &\checkmark  &\texttimes &\checkmark \\
Cavity Flow\cite{Lid}   &\texttimes & \checkmark & \texttimes & 1000 & 60 & 1 & 1 &\texttimes &\checkmark  &\texttimes &\checkmark\\
Granular flow \cite{boundary} &\texttimes & \texttimes & \checkmark  & 365 & 250 & 2 & 1 &\checkmark & \texttimes  &\checkmark &\texttimes\\
Kubric MOVi-A\cite{Kubric}  &\texttimes & \texttimes & \checkmark  & -- & -- & 1 & 3 &\checkmark & \texttimes  &\checkmark &\texttimes\\
ROPE \cite{rope} &\texttimes &\checkmark &\texttimes  & 10000 & 160 & 1 & 1 &\checkmark & \texttimes  &\checkmark &\texttimes\\
\hline
3 mode system \cite{3}  &\texttimes &\checkmark &\texttimes   & 10500 & 180 & 1 & 3  &\checkmark & \texttimes  &\checkmark &\texttimes\\
Bouncing ball \cite{3}  &\checkmark &\texttimes  &\texttimes   & 1100 & 100 & 1 & 1  &\checkmark & \texttimes  &\checkmark &\texttimes\\
Simulated Cloth \cite{cloth} &\texttimes &\checkmark  &\checkmark   & 1050 & 160 & 1 & 1 & \texttimes &\checkmark   &\checkmark &\texttimes\\
Deformable Plate \cite{sensor} &\texttimes &\texttimes  &\checkmark   & 675 & 50 & 1 & 1 &\checkmark &\texttimes    &\checkmark &\texttimes\\
\hline
\textbf{MBDS(ours)} & \textbf{\checkmark} &\textbf{\checkmark} &\textbf{\checkmark} & \textbf{15000} & \textbf{200} & \textbf{8} & \textbf{32(Ave)} &\textbf{\checkmark} & \textbf{\checkmark }&\textbf{\checkmark} &\textbf{\checkmark}  \\
 \hline
\bottomrule
\end{tabular}
}
\end{table*}

Most of the aforementioned methods for simulating physical systems are based on GNNs\cite{GNS}, which restructure the studied objects into graph representations. One category of methods utilizes point cloud techniques to transform objects into point clouds \cite{sensor}, which are then converted into graphs. Another subset relies on finite element methods \cite{Mesh/iclr/PfaffFSB21} to grid objects before transforming them into graph-structured data. These approaches have demonstrated favorable outcomes in analytical tasks such as object deformation \cite{sensor}, fluid states \cite{Lid}, and rigid body collisions \cite{rigidcube}. 

However, there is still a significant gap between the theoretical advancements and practical applications of current neural network-based methods for simulating physical systems in real-world scenarios. This gap can be attributed to the inability of existing simulation methods to fully capture the complexity of real-world conditions. Many existing methods primarily focus on analyzing the interactions between individual objects or pairs, which does not align with the practical industrial applications of object dynamics simulation. 

In actual industrial scenarios, simulation algorithms often need to analyze the interactions among various complex mechanical structures. This is a challenge that current datasets struggle to simulate. This limitation results in the current neural network-based methods for simulating physical systems remaining largely theoretical and unable to make significant strides in real-world industrial applications. To bridge the gap between theoretical research and practical industrial applications, there is a pressing need for more complex multi-body dynamics simulation datasets. These datasets should encompass more intricate mechanical structures and involve scenarios with three or more objects, incorporating rotational dynamics, collisions, and various other complexities. This expansion is crucial for broadening the applicability of these methods and enabling them to address the complexities of real-world industrial dynamics.

\begin{figure}[t]
    \centering
	\begin{minipage}{1\textwidth} 
			\includegraphics[width=\textwidth]{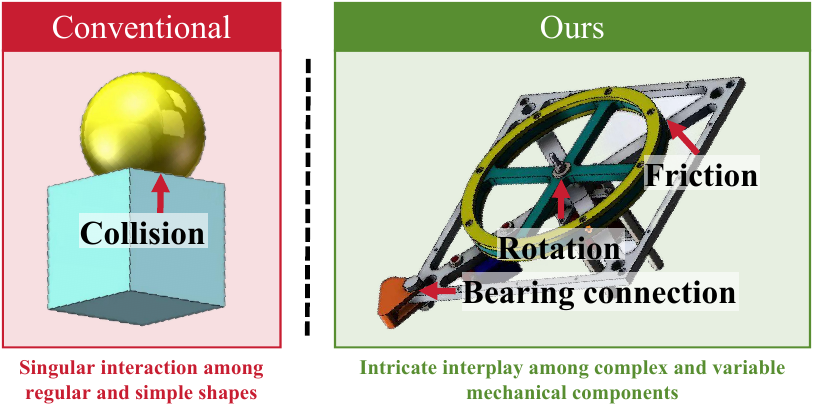} \\
	\end{minipage}

	\caption{A comparison between our dataset and conventional datasets}
	\label{fig:element}
     \vskip -0.1in
\end{figure}
To address the aforementioned issue, we propose the \textit{ \textbf{M}ulti-\textbf{B}ody \textbf{D}ynamics \textbf{S}imulation} (MBDS) dataset, to provide a simulation dataset with a broader and more complex range of scenarios that closely aligns with real-world conditions for the analysis of neural network-based methods in physical systems. In Table~\ref{tab:comparision}, we present a comparison between our dataset and other commonly used datasets. The MBDS dataset comprises 150,000 motion trajectories along with more intricate mechanical linkage structures, facilitating simulations of even more complex multi-body dynamics scenarios. When constructing the data for the MBDS database, we incorporated complex mechanical structures that closely mirror those found in real industrial settings, as opposed to employing simplistic shapes like spheres or cubes. Figure~\ref{fig:element} provides a specific example to compare the physical scenes modeled by our dataset with those modeled by other existing datasets. The main advantages of the MBDS dataset can be summarized as follows: 
\begin{itemize}
\item The MBDS dataset closely aligns with real-world engineering scenarios, as it is designed based on complex mechanical structures encountered in practical engineering contexts. 
\item The MBDS dataset offers a larger volume of data compared to existing relevant datasets, with the highest or a substantial quantity of trajectories, time steps, scenes, and object counts. 
\item The MBDS dataset encompasses a broader range of scenarios, including various force applications and motion situations, enabling a more comprehensive validation of related methodologies.
\end{itemize}

By leveraging our proposed MBDS dataset, a series of analytical experiments are conducted to facilitate an in-depth analysis of the experimental results. Our research outcomes elucidate two critical facets of trajectory prediction within our datasets. As the temporal horizon of the prediction extends, particularly in the context of complex multi-body dynamics datasets, we not only observe an incremental increase in predictive error but also discern a propensity for this error to escalate exponentially. This trend underscores a pronounced lack of robustness in the prevailing modeling framework. Furthermore, it becomes evident that numerous models exhibit proficiency solely within a specific scenario, and this proficiency experiences a marked decline when the scenario undergoes alteration. Even minor adjustments within the same scenario, such as variations in speed, lead to inaccurate predictions, signaling a deficiency in the generalizability of existing models.

\section{Related Works}
\subsection{Graph Neural Networks Simulators}
In recent years, there has been a shift towards more machine learning-centric methodologies in physical simulators to better support general physical dynamics\cite{UMME,litt,2007position,flexible}. Among these, methods based on GNNs stand out as particularly effective\cite{NMP}. GNNs adeptly map an input graph to an equivalent structure output graph, yet with potentially different attributes for nodes, edges, and graphs-level, and can be trained to learn a form of message-passing, where latent information is propagated between nodes via the edges\cite{gn2018,2008graph}. Owing to the capability of GNNs to effectively handle message passing between nodes and representing entities and their relations with graphs, and compute their interactions, numerous works have employed Graph Neural Networks as physical simulators for dynamical systems. More and more GNNs and their variants\cite{cloth, BMS}, such as Interaction Networks\cite{016interaction}, have been adeptly learned to simulate rigid bodies, mass-spring systems, n-body problems, and robotic control systems\cite{GNS, yl17}, as well as non-physical systems like multi-agent dynamics, algorithm executions, and other dynamic graph settings. 
\subsection{Datasets for Graph Neural Networks Simulators}  
\begin{itemize}

\item \textbf{Simplified Scenarios:} Initially, GNS-related datasets primarily focused on simple and controlled scenarios. These datasets typically featured basic physical systems, such as single-body dynamics\cite{GNS, MD17} or simplified multi-particle problems like pendulums\cite{LGNN,3} and springs\cite{HGNN}. While useful for initial explorations in the field, these datasets lacked the complexity needed to fully test and develop advanced GNS models.
\item \textbf{Advanced Physical Interactions:} As GNS technology progressed, there was a shift towards incorporating more complex physical interactions. Datasets began to include dynamic interactions between multiple entities, encompassing scenarios like fluid dynamicsp\cite{Lid, boundary} and deformable materials. These datasets provided a richer environment for testing GNS capabilities, though they often required significant computational resources due to increased complexity.
\item \textbf{Integration of Real-World Scenarios:} Acknowledging the demand for simulations that better mirror actual conditions, recent datasets have begun incorporating scenarios that closely resemble real-world environments\cite{bear2021physion}, but this approach, often relying on sensors or computer vision\cite{sensor}, can be marred by issues such as device jitter, electromagnetic interference, and deviations during signal transmission.

\end{itemize}
\section{Dataset Construction}
The design of the MBDS dataset necessitates meticulous consideration of three critical factors: the data collection strategy, adherence to physical laws, and ensuring both ease and consistency in its utilization. The entire design process is chiefly focused on rectifying two notable deficiencies in previous datasets: the information provided is overly simplistic in terms of speed and position, and it notably lacks consideration of external force sources beyond gravity. %Only by addressing the aforementioned issues can the constructed dataset effectively serve as a good benchmark.
Addressing the multifaceted nature of multi-body dynamical systems and the requirement for dataset universality, the construction phase is inherently time-intensive and complex. In the following sections, we meticulously outline the data collection methodology and explore the significant advantages it offers.

\subsection{Data Collection}\label{AA}

Given the current limitations of computer vision technology, which often leads to information noise and resultant errors, and considering the inconvenience and high costs associated with sensor-based data collection, we opt to perform scenario modeling and data collection through the application of physical laws of multibody dynamics and high-precision calculations. The MBDS include 3D scenarios such as a four-wheel ParticleCar, a SimpleCubli, a 3-Ladder, a 2-Ladder, a 6-Pendulum, and a 5-Pendulum; 2D scenarios such as the simple projectile motion of a BallDrop and 1D models such as a BallScrol. These eight are simple yet very representative scenarios. Each scenario consists of \textbf{15000} trajectories with \textbf{200} time steps. The dataset is divided into two parts: one for training, containing 13500 trajectories, and another for evaluation, containing 1500 trajectories. We instantiate every particle, using the parameters listed in Table~\ref{tab:mujoco} and we will introduce the processing details in the following: 
\begin{table}[h]
\caption{Particle Parameters}
\label{tab:mujoco}
\begin{center}
\begin{tabular}{c|c|c|c}
\hline
\textbf{Constant} & \textbf{\textit{Symbol}} & \textbf{\textit{Value}} & \textbf{\textit{Units}} \\
\hline
mass & \( m \) & 1 & kg \\
\hline
inertia & \( I \) & \( 6.17 \times 10^{-4} \) & kg\(\cdot\)m\(^2\) \\
\hline
force & \( F \) & (various) & N \\
\hline
gravity & \( g \) & 9.81 & m/s\(^2\) \\
\hline
friction coefficient & \( \mu \) & \(4.3\times 10^{-1}\) & (none) \\
\hline

damping ratio & \( \zeta \) & \(2.2 \times 10^{-2}\) & (none) \\
\hline
time-step & \( \Delta t \) & \( 2 \times 10^{-2} \) & s \\
\hline
\end{tabular}
\label{tab:die_roll_parameters}
   
\end{center}
 \vskip -0.4in
\end{table}

\begin{itemize}
\item \textbf{Step 1:} Start by creating a foundational particle, and then develop subsequent particles based on this initial model. Each particle is meticulously simulated, incorporating realistic physical attributes like mass and friction. Most crucially, constraints are introduced between particles. In the case of rigid bodies, these constraints maintain a consistent distance between them, thereby creating a complex multi-body system where each particle not only is influenced by but also impacts its neighboring particles.
\item \textbf{Step 2:} To enhance the realism of our simulations, we introduce noise interference at each step. This involves applying small, random perturbations to mimic real-world disturbances such as air resistance and friction, which are often challenging to measure accurately. By incorporating these elements, our approach more effectively mirrors the complexities and unpredictability of real-world scenarios.
\item \textbf{Step 3:} Our dataset represents a significant enhancement in the diversity of force applications on objects, markedly extending beyond the traditional focus on gravitational forces. This encompasses an array of force scenarios: 
\textbf{\romannumeral1)} Dynamics scenarios, including a four-wheel ParticleCar and a SimpleCubli model, demonstrate systems in motion under variable forces and velocities; \textbf{\romannumeral2)} Rope ladder scenarios, exemplified by 3-Ladder and 2-Ladder configurations, illustrate the behavior of flexible structures with one end anchored and the other end responding to external forces, such as wind; \textbf{\romannumeral3)} Pendulum scenarios, namely 6-Pendulum and 5-Pendulum setups, provide fundamental test cases for physical modeling; \textbf{\romannumeral4)} Additionally, other scenarios, like BallDrop and BallScrol, focus on object trajectories predominantly influenced by gravity. Each example serves to highlight the broad applicability and enhanced realism of our dataset in simulating complex physical interactions.
\end{itemize}

% \begin{figure}[htbp]
% \centerline{\includegraphics{car.jpg}
% \caption{Example of a figure caption.}
% \label{fig}
% \end{figure}

\subsection{Statistics and Analysis}
This section conducts some necessary statistics and analysis to gain a better understanding of the proposed dataset MBDS. From the statistics, we observe the following advantages:
\begin{itemize}
\item \textbf{Purity:} Compared to other particle datasets derived through image processing, which invariably introduces noise during the conversion of images to particles\cite{Physion}, or datasets obtained from real-world sensors that may exhibit biases in signal transmission and conversion\cite{sensor}, our dataset stands out in its purity. By meticulously following the physical laws governing multi-body dynamics and accounting for a myriad of disturbance factors encountered in real-world situations, we successfully create the most direct and pure physical dataset available. This approach ensures a higher level of accuracy and reliability in our data, pivotal for advancing research in this field.
\item \textbf{Versatile and Malleable Forms:} In our dataset, careful consideration has been given to practical physical implications, as well as the generalizability strengths and weaknesses of GNS models, even within identical scenarios. This led us to create datasets encompassing a range of different speeds and forces. Our experiments show that GNS models predictions tend to decrease in accuracy at higher velocities. For instance, in our ParticleCar scenario, we divide the speed ranges for testing into three intervals: 10-30, 30-70, and 70-100. In this setup, the speeds of the four wheels are not uniform, varying within these intervals, to simulate the effects of steering. Similarly, forces are calibrated based on the scenario, with three specific intervals set to more rigorously evaluate the models' performance under varied conditions. This structured approach enables a more effective assessment of the models' capabilities. 

\item \textbf{Scalability:} In our comprehensive analysis of existing datasets, we observe a predominant reliance on fixed data formats such as .h5 or .npz. While these formats are not inherently problematic, the primary issue arises from their often inconsistent and non-scalable nature in typical dataset structures, which hinders effective scenario extension and adaptation. To mitigate these limitations, our dataset is made available in two widely-used formats, alongside providing the rawest form of data in .csv files, thereby facilitating seamless integration and comparison with diverse datasets. Further, in our commitment to collaborative advancement, we will release the dataset generation code to the public domain. This move is intended to empower fellow researchers in the field, enabling them to tailor and expand the dataset in alignment with their unique research objectives, thus fostering innovation and progress in the realm of GNS.

\item \textbf{Benchmark:} In conclusion, our dataset is meticulously designed to closely mirror real-world scenarios, offering a broad and relevant platform for evaluating various GNS models. It is comprehensive, incorporating a diverse range of scenarios from complex 3D environments to fundamental 2D and 1D setups. This diversity ensures a robust and flexible evaluation framework, establishing our dataset as an ideal benchmark for thoroughly examining current models. Furthermore, this comprehensive nature not only allows for rigorous testing but also facilitates a detailed analysis of each network architecture, identifying its strengths and weaknesses. Consequently, our dataset serves as an invaluable benchmark in guiding the enhancement and innovation of network designs.

\end{itemize}
% \begin{figure}[t]
%  \centering
%  \begin{minipage}[t]{0.9\linewidth}
%         \centering
%         \includegraphics[width=1.0\linewidth]{result_fig/ParticleCar.pdf}
%  \end{minipage}%
%  \caption{The rollout-MSE curves on ParticleCar.}
%  \label{fig:car}
% \end{figure}

% \begin{figure}[t]
% % \vspace{-4pt}
%  \centering
%  \begin{minipage}[t]{0.9\linewidth}
%         \centering
%         \includegraphics[width=1.0\linewidth]{result_fig/SimpleCubli.pdf}
%  \end{minipage}%
%  \caption{The rollout-MSE curves on SimpleCubli.}
%  \label{fig:cubli}
% \end{figure}

% \begin{figure}[t]
%  \centering
%  \begin{minipage}[t]{0.9\linewidth}
%         \centering
%         \includegraphics[width=1.0\linewidth]{result_fig/6—Pendulum.pdf}
%  \end{minipage}%
%  \caption{The rollout-MSE curves on 6—Pendulum.}
%  \label{fig:pendulum}
% \end{figure}
% \begin{figure}[t]
% % \vspace{-4pt}
%  \centering
%  \begin{minipage}[t]{0.9\linewidth}
%         \centering
%         \includegraphics[width=1.0\linewidth]{result_fig/3-Ladder.pdf}
%  \end{minipage}%
%  \caption{The rollout-MSE curves on 3-Ladder.}
%  \label{fig:ladder}
% \end{figure}

\section{Evaluation and Benchmarks}
In this section, we present a standardized evaluation pipeline utilizing our datasets for object pose detection and trajectory prediction. Employing the latest Graph Network Simulator (GNS) models, we benchmark the performance across various scenarios and provide a detailed discussion of the results.

%In this section, we introduce a standardized evaluation pipeline using our datasets for object pose detection and trajectory prediction. We use the latest models of GNS to benchmark the performance of each scenario and discuss their results.
% \input{table/Car_mse}
% \input{table/pendulum}
% \input{table/cubli}
% \input{table/ladder}

\subsection{Experimental Setting and Metrics}\label{AA}
\begin{itemize}
\item \textbf{Settings:} In our methodology, we split the entire dataset into a training set and a validation set using a 9:1 ratio. Each model underwent training 10 times, with each training session spanning 200 epochs. We record the average performance on the validation set to ensure a robust evaluation. To maintain fairness in our comparisons, we keep all hyperparameters consistent across models without any additional tuning. Specifically, all MLPs are initialized with three projection layers and a hidden dimension of 200. For optimization, we employ an Adam optimizer with an initial learning rate of 0.0001 and betas set at (0.9, 0.999). Additionally, we utilized a Plateau scheduler, applying a patience of three epochs and a decay factor of 0.8.

\item \textbf{Metrics:} We adhere to the experimental setup outlined above and adopt positional error as our evaluation standard. To introduce the calculation of Mean Squared Error(MSE) briefly, it is computed by taking the mean of the squared differences between the predicted values and the actual values. This method is highly effective in evaluating the accuracy of the discrepancy between rollout and ground truth, serving as a commonly used metric in trajectory prediction research.

\end{itemize}

\subsection{Baseline models}\label{AA}

To demonstrate the superiority of our dataset, we conducted experiments based on our datasets using various GNS models, which include: 
\begin{itemize}
\item \textbf{HGNN:}Learning the Dynamics of Physical Systems with Hamiltonian Graph Neural Networks. (Suresh Bishnoi et al. 2023)
\item \textbf{LGNN:}Learning the Dynamics of Particle-based Systems with Lagrangian Graph Neural Networks. (Suresh Bishnoi et al. 2023)
\item \textbf{SGNN:}Learning Physical Dynamics with Subequivariant Graph Neural Networks. (Jiaqi Han et al. 2022)
\item \textbf{GNS:}Learning to Simulate Complex Physics with Graph Networks. (Alvaro Sanchez-Gonzalez et al. 2021)

\end{itemize}
\begin{figure}[ht]
	\centering
  	\subfigure[ParticleCar]{
		\begin{minipage}{0.3\textwidth} 
			\includegraphics[width=\textwidth]{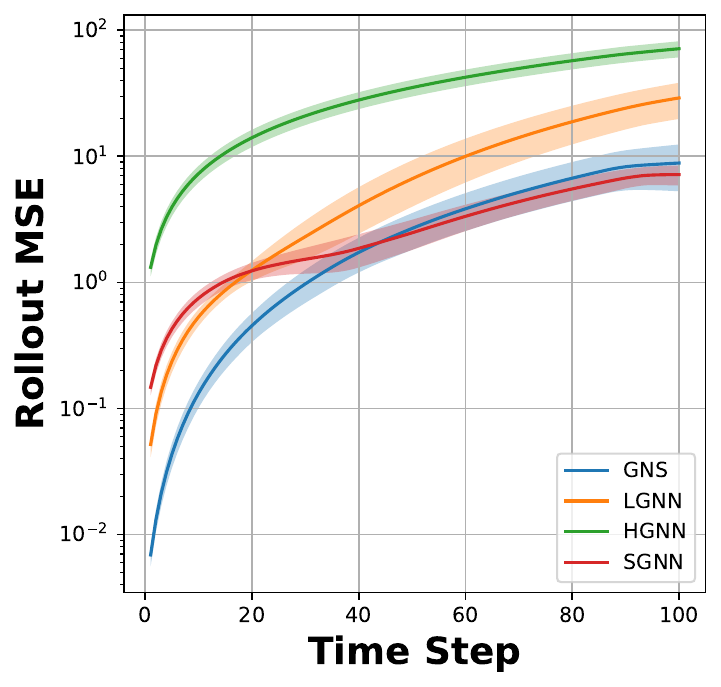} \\
		\end{minipage} 
	} \hspace{1mm}
	\subfigure[6-Pendulum]{
		\begin{minipage}{0.3\textwidth} 
			\includegraphics[width=\textwidth]{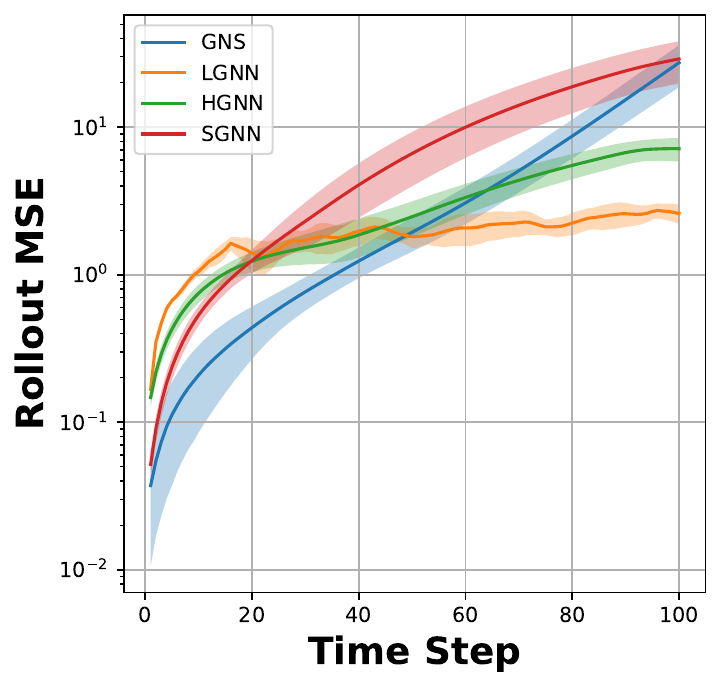} \\
		\end{minipage} 
	}
 	\subfigure[SimpleCubli]{
		\begin{minipage}{0.3\textwidth} 
			\includegraphics[width=\textwidth]{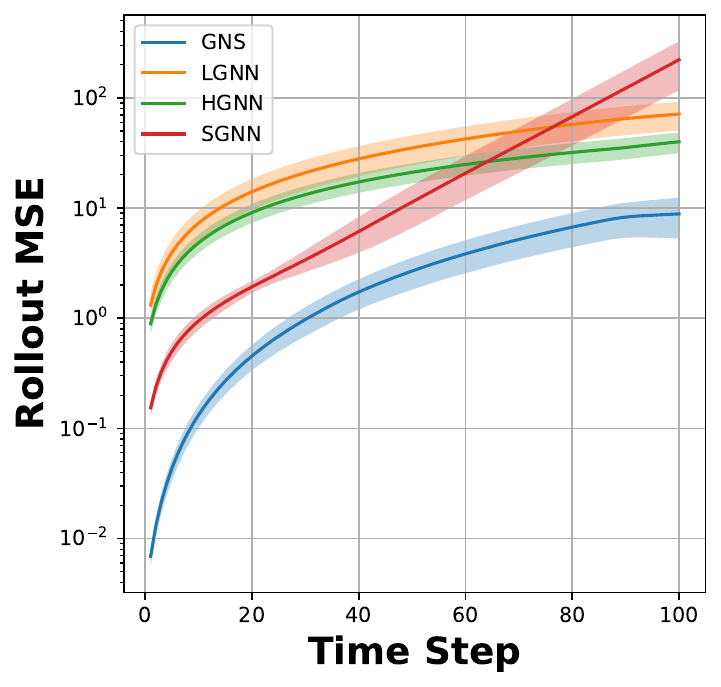} \\
		\end{minipage} 
	}

   	\subfigure[3-Ladder]{
		\begin{minipage}{0.3\textwidth} 
			\includegraphics[width=\textwidth]{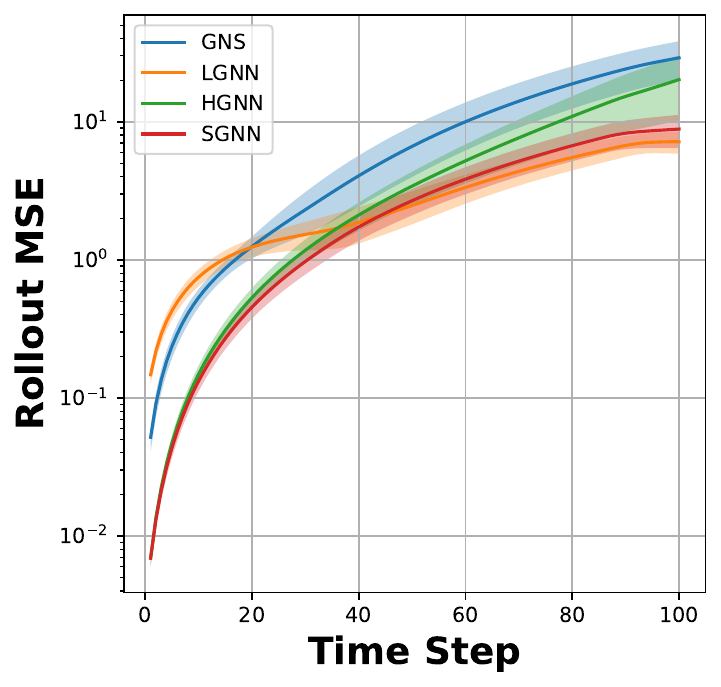} \\
		\end{minipage} 
	} \hspace{1mm}
	\subfigure[BallDrop]{
		\begin{minipage}{0.3\textwidth} 
			\includegraphics[width=\textwidth]{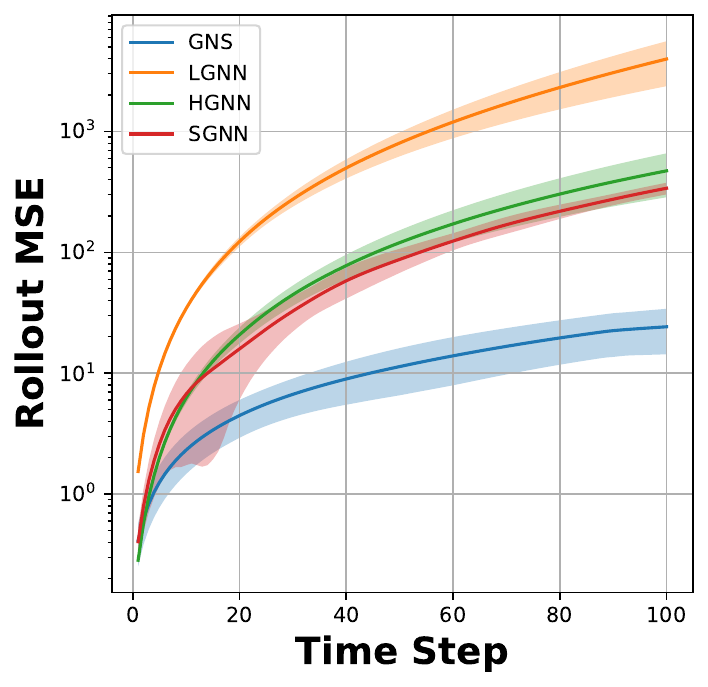} \\
		\end{minipage} 
	}
 	\subfigure[BallScroll]{
		\begin{minipage}{0.3\textwidth} 
			\includegraphics[width=\textwidth]{result_pdf/car_loss_plot.pdf} \\
		\end{minipage} 
	}

	\caption{Experimental results on precision evaluation.}
	\label{fig:curves-all}
\end{figure}
\subsection{Results and discussion}\label{AA}
We present the rollout MSE for t = 1 and t = 50 on the MBDS dataset in Table~\ref{tab:result:1} and Table~\ref{tab:result:50}. In these analyses, we use the basic GNS model as a baseline and compare it with the latest variants of GNS network architectures. This comparison allows us to glean more detailed insights into the four models from the tabulated data. Figure \ref{fig:curves-all} illustrates the curve plot of our experimental results on precision evaluation. It is evident that our dataset serves effectively as a benchmark and can be utilized as a tool for analyzing the shortcomings present in current models. This underscores the effectiveness of our proposed MBDS dataset. Based on these results as observed in the Rollout-MSE curves for some scenarios, we will subsequently conduct an in-depth analysis to discern the strengths and weaknesses of each model, particularly in the context of their performance against the established baseline. (In some result tables, \textcolor{green}{green} indicates improvement, and \textcolor{red}{red} signifies inferiority.
)
\begin{table*}[t]
\renewcommand\arraystretch{1.5}
\caption{Rollout MSE $(\times 10^{-2})$ on MBDS when \( t \)=1}
\label{tab:result:1}
\centering
\scalebox{0.77}{
\begin{tabular}{l c c c c c c c c }
\toprule
\ & \textbf{ParticleCar} & \textbf{6-pundulum}  & \textbf{5-pendulum} & \textbf{3-ladder} &\textbf{2-ladder} & \textbf{SimpleCubli}  & \textbf{BallDrop} & \textbf{BallScroll} \\ 
\bottomrule
\rowcolor{gray!20} \textbf{GNS}    &0.72$ \pm$0.60   &1.44$ \pm$ 0.34    &0.39$ \pm$ 0.23    &0.41$\pm$0.04  &0.29$ \pm$0.16    &2.23$ \pm$0.14 &0.13$ \pm$0.01   &0.17 $\pm$ 0.01     \\
\toprule
\textbf{SGNN}   &1.93 $\pm$ 0.28     &1.32$\pm$ 0.65   &1.24$\pm$1.92   &0.83$\pm$0.19   &0.54$\pm$0.10   &1.34$\pm$0.44  &0.45$\pm$0.09   &0.23$\pm$0.02   \\
\textbf{$\Delta$} & \textcolor{red}{1.21} $\uparrow$ & \textcolor{green}{-0.12} $\downarrow$ & \textcolor{red}{0.85} $\uparrow$ & \textcolor{red}{0.42} $\uparrow$ & \textcolor{red}{0.25} $\uparrow$ & \textcolor{green}{-0.89} $\downarrow$ & \textcolor{red}{0.32} $\uparrow$ & \textcolor{red}{0.06} $\uparrow$ \\
\hline
\textbf{HGNN}    &2.89$\pm$1.12    &1.50$\pm$1.27   &1.31$\pm$1.74   &0.95$\pm$0.27   &0.78$\pm$0.20   &2.34$\pm$0.31   &0.51$\pm$0.23  &0.15$\pm$0.12   \\
\textbf{$\Delta$} & \textcolor{red}{2.17} $\uparrow$ & \textcolor{red}{0.06} $\uparrow$ & \textcolor{red}{0.92} $\uparrow$ & \textcolor{red}{0.54} $\uparrow$ & \textcolor{red}{0.49} $\uparrow$ & \textcolor{red}{0.11} $\uparrow$ & \textcolor{green}{-0.30} $\downarrow$ & \textcolor{green}{-0.02} $\downarrow$ \\
\hline
\textbf{LGNN}   &1.74$\pm$0.42   &1.58$\pm$1.04     &1.11$\pm$0.39   &4.36$\pm$1.33  &0.45$\pm$0.23   &3.20$\pm$0.52    &0.25$\pm$0.14 & /0.14$\pm$0.03   \\
\textbf{$\Delta$} & \textcolor{red}{1.02} $\uparrow$ & \textcolor{green}{-0.04} $\downarrow$ & \textcolor{red}{0.72} $\uparrow$ & \textcolor{red}{3.95} $\uparrow$ & \textcolor{red}{0.16} $\uparrow$ & \textcolor{red}{0.97} $\uparrow$ & \textcolor{red}{0.12} $\uparrow$ & \textcolor{green}{-0.03} $\downarrow$ \\
\toprule
\end{tabular}
}
    \vskip -0.1in
\end{table*}
\begin{table*}[h]
\renewcommand\arraystretch{1.5}

\caption{Rollout MSE on MBD when \( t \)=50}
\label{tab:result:50}
\centering
\scalebox{0.78}{
\begin{tabular}{l c c c c c c c c }
\toprule
\ & \textbf{ParticleCar} & \textbf{6-pundulum}  & \textbf{5-pendulum} & \textbf{3-ladder} &\textbf{2-ladder} & \textbf{SimpleCubli}  & \textbf{BallDrop} & \textbf{BallScroll} \\ 
\bottomrule
\rowcolor{gray!20}\textbf{GNS}    &2.98$\pm$0.24   &1.94$\pm$0.29    &1.75$\pm$1.43    &6.75$\pm$2.94  &5.64$\pm$0.87    &2.34$\pm$0.34 &1.75$\pm$1.43   &0.37$\pm$0.21     \\
\toprule
\textbf{SGNN}   &2.56$\pm$0.48     &8.32$\pm$2.65   &6.24$\pm$1.92   &1.05$\pm$0.79   &1.04$\pm$0.49   &5.64$\pm$1.54  &2.65$\pm$1.43   &0.53$\pm$0.12   \\
\textbf{$\Delta$} & \textcolor{green}{-0.42} $\downarrow$ & \textcolor{red}{6.38} $\uparrow$ & \textcolor{red}{4.49} $\uparrow$ & \textcolor{green}{-5.70} $\downarrow$ & \textcolor{green}{-4.60} $\downarrow$ & \textcolor{red}{3.30} $\uparrow$ & \textcolor{red}{0.90} $\uparrow$ & \textcolor{red}{0.16} $\uparrow$ \\
\hline
\textbf{HGNN}    &29.13$\pm$3.72    &2.73$\pm$1.27   &2.67$\pm$0.34   &3.05$\pm$1.02   &2.94$\pm$0.86  &15.24$\pm$1.82   &1.45$\pm$1.43  &0.74$\pm$0.02   \\
\textbf{$\Delta$} & \textcolor{red}{26.15} $\uparrow$ & \textcolor{red}{0.79} $\uparrow$ & \textcolor{red}{0.92} $\uparrow$ & \textcolor{green}{-3.70} $\downarrow$ & \textcolor{green}{-2.70} $\downarrow$ & \textcolor{red}{12.90} $\uparrow$ & \textcolor{green}{-0.30} $\downarrow$ & \textcolor{red}{0.37} $\uparrow$ \\
\hline
\textbf{LGNN}   &8.87$\pm$2.43   &1.90$\pm$1.18     &1.82$\pm$0.56   &2.17$\pm$0.75  &1.74$\pm$0.37   &21.97$\pm$1.49    &4.75$\pm$1.43 &0.32$\pm$0.01   \\
\textbf{$\Delta$} & \textcolor{red}{5.89} $\uparrow$ & \textcolor{green}{-0.04} $\downarrow$ & \textcolor{red}{0.07} $\uparrow$ & \textcolor{green}{-4.58} $\downarrow$ & \textcolor{green}{-3.90} $\downarrow$ & \textcolor{red}{19.63} $\uparrow$ & \textcolor{red}{3.00} $\uparrow$ & \textcolor{green}{-0.05} $\downarrow$ \\
\toprule
\end{tabular}
}
    \vskip -0.1in
\end{table*}

\begin{table*}[t]
    \renewcommand{\arraystretch}{1.2}
    \setlength{\tabcolsep}{8pt}

\centering
    \caption{Comparative experiment results. \textbf{Bold} indicates the method with optimal performance. \underline{Underline} denotes the method with second-best performance. Some standard deviations are marked as 0.00 due to being too small to effectively represent.}
    \label{tab:main_result}
    \scalebox{0.85}{     

    \begin{tabular}{c|l|ccccc}
        \hline
               \multicolumn{2}{c|}{Senario} & \multirow{2}{*}{ParticleCar} &  \multirow{2}{*}{6-Pendulum} & \multirow{2}{*}{SimpleCubli} &\multirow{2}{*}{3-Ladder}&\multirow{2}{*}{BallDrop}\\
        \cline{1-2}
        Time & Methods   \\
        \hline
        \multirow{4}{*}{\rotatebox[origin=c]{90}{t=10}} 
        \scriptsize\multirow{4}{*}{\rotatebox[origin=c]{90}{(MSE $\times 10^{0}$)}}
         & GNS  & \textbf{0.11$\pm$0.06} & \textbf{0.21$\pm$0.05} & \underline{0.57$\pm$0.05}  & 0.77$\pm$0.01 & 0.23$\pm$0.02 \\
         & SGNN & 0.81$\pm$0.08 & \underline{0.45$\pm$0.08} & \textbf{0.43$\pm$0.37} & \textbf{0.14$\pm$0.01} & \underline{0.33$\pm$0.08} \\
         & HGNN & 3.73$\pm$0.44 & 0.61$\pm$0.07 & 6.69$\pm$1.03 & \underline{0.21$\pm$0.03} &0.17$\pm$0.06 \\
         & LGNN & \underline{0.41}$\pm$0.23 & 0.87$\pm$0.26 & 4.32$\pm$0.98 & 0.81$\pm$0.32 & \textbf{0.16$\pm$0.03} \\
        % & MDGS (Ours)  & \textbf{0.84$\pm$0.28}  & \textbf{1.76$\pm$0.63}    & \textbf{2.08$\pm$0.24}   &  \textbf{0.84$\pm$0.04}   \\
        \hline
        \multirow{4}{*}{\rotatebox[origin=c]{90}{t=20}} 
        \scriptsize\multirow{4}{*}{\rotatebox[origin=c]{90}{(MSE $\times 10^{0}$)}}
             & GNS  & \textbf{0.38$\pm$0.07}     & \textbf{0.39$\pm$0.15}     &\textbf{1.27$\pm$0.14}     & \underline{1.01$\pm$0.14} & \textbf{0.26$\pm$0.01} \\
             & SGNN & 1.05$\pm$0.41              & \underline{1.29$\pm$0.35}  & \underline{2.34$\pm$1.27} & \textbf{0.62$\pm$0.03} &0.41$\pm$0.12 \\
             & HGNN & 11.97$\pm$2.44               & 1.33$\pm$0.28              & 9.37$\pm$2.06             & 0.72$\pm$0.26 & \underline{0.35$\pm$0.03} \\
             & LGNN & \underline{1.03$\pm$0.64}     & 1.32$\pm$0.54              & 12.10$\pm$1.34          & 1.03$\pm$0.17 & 0.39$\pm$0.05 \\
        % & MDGS (Ours)  & \textbf{0.84$\pm$0.28}  & \textbf{1.76$\pm$0.63}    & \textbf{2.08$\pm$0.24}   &  \textbf{0.84$\pm$0.04}   \\
        \hline
                \multirow{4}{*}{\rotatebox[origin=c]{90}{t=40}} 
                \scriptsize\multirow{4}{*}{\rotatebox[origin=c]{90}{(MSE $\times 10^{1}$)}}
         & GNS  & \underline{0.15$\pm$0.01} & \textbf{0.11$\pm$0.02} & \textbf{0.19$\pm$0.02} & 0.37$\pm$0.01 & \textbf{0.03$\pm$0.00} \\
         & SGNN & \textbf{0.12$\pm$0.02} & 0.50$\pm$0.14 & \underline{0.83$\pm$0.29} & \textbf{0.14$\pm$0.03} & 0.08$\pm$0.01 \\
         & HGNN & 2.31$\pm$0.40 & \underline{0.16$\pm$0.06} & 1.90$\pm$0.24 & 0.17$\pm$0.05 & 0.07$\pm$0.02 \\
         & LGNN & 1.53$\pm$0.26 & 0.21$\pm$0.07 & 2.43$\pm$0.64 & \underline{0.15$\pm$0.09} & \underline{0.04$\pm$0.00} \\
        % & MDGS (Ours)  & \textbf{0.17$\pm$0.08} & \textbf{0.24 $\pm$0.06}    & \textbf{0.31$\pm$0.03}   & \textbf{0.15$\pm$0.04}   \\
        \hline

                        \multirow{4}{*}{\rotatebox[origin=c]{90}{t=100}} 
                    \scriptsize\multirow{4}{*}{\rotatebox[origin=c]{90}{(MSE $\times 10^{1}$)}} 
         & GNS  & \underline{0.99$\pm$0.07} & 2.16$\pm$0.52 & \textbf{0.94$\pm$0.19} & 1.24$\pm$0.01 & \textbf{0.21$\pm$0.08} \\
         & SGNN & \textbf{0.93$\pm$0.27} & 2.64$\pm$0.44 & 11.98$\pm$1.07 & \underline{0.93$\pm$0.01} & \underline{0.34$\pm$0.23} \\
         & HGNN & 8.64$\pm$0.51 & \underline{0.78$\pm$0.39} & \underline{3.47$\pm$0.76} & 1.05$\pm$0.06 & 0.45$\pm$0.11 \\
         & LGNN & 2.34$\pm$0.43 & \textbf{0.28$\pm$0.08} & 4.14$\pm$0.89 & \textbf{0.88$\pm$0.17} & 0.64$\pm$0.06 \\
        % & MDGS (Ours)  & \textbf{0.05$\pm$0.01}   & \textbf{0.06$\pm$0.01}    & \textbf{0.09$\pm$0.01}  & \textbf{0.04$\pm$0.01} \\
        \hline
    
    \end{tabular}
}
    \vskip -0.1in

\end{table*}

\begin{enumerate}

    \item In our comparative analysis focused on the ParticleCar scenario across four models, the results, as detailed in Table~\ref{tab:main_result}, reveal that SGNN and HGNN underperform in comparison to the basic GNS model. Notably, LGNN exhibits the poorest performance. This disparity in model efficacy is particularly pronounced in scenarios where each particle exhibits distinct intrinsic dynamics. These findings suggest that the SGNN, HGNN, and LGNN while predicting trajectories, do not adequately account for situations where particles are subjected to forces beyond mere gravity. This lack of comprehensive force consideration appears to be a significant factor in the reduced accuracy of these models compared to the basic GNS.
    
    \item Based on the 4 models tested in the 6-Pendulum scenario, a common setup involves applying an initial velocity or disturbance, followed by pendulum motion under the influence of gravity. The results in Table~\ref{tab:main_result} indicate that due to SGNN's consideration of the consistent vertical direction of gravity, HGNN's account for the constancy of external forces over time in line with the conservation of Hamiltonian, and Lagrangian Law of Conservation of Kinetic and Potential Energy for LGNN, their performance is markedly superior to that of GNS.
    
    \item In the scenario of the SimpleCubli, where a model is thrown from a certain height and influenced by gravity, additionally equipped with three flywheels attempting to achieve Cubli's body balance through torque changes, the experimental results, as shown in Table~\ref{tab:main_result} demonstrate that in such a scenario with mixed dynamics, the basic GNS performs the best. Given that this scenario does not conform to Hamiltonian mechanics and Lagrangian Law, it is speculated that HGNN and LGNN perform worse than the basic GNS. SGNN, taking into account the symmetry of the system, but due to the random dropping scenario, does not satisfy the symmetry, resulting in the worst performance.
    
    \item In the rope ladder scenario, modeled to simulate a helicopter rope ladder with one end fixed and the other swaying in the wind, we observed a dominant influence of gravity, with wind force serving as a minor perturbative factor. As shown in Table~\ref{tab:main_result}, SGNN demonstrates superior predictive performance, likely due to its effective integration of gravity as a core element in the model. Meanwhile, HGNN and LGNN exhibit moderate performance. In contrast, the commonly used GNS, lacking this specific focus on gravity, underperforms in this scenario.
    
    \item In the simplest scenario, BallDrop, where a ball naturally falls, we observe in Table~\ref{tab:main_result} that the performance differences among the four methods are minimal, demonstrating that existing models are capable of adequately completing the task of simple trajectory prediction. Interestingly, as time steps progress, the most basic model, GNS, exhibits commendable performance, primarily because it makes positional judgments based solely on fundamental positional information. Conversely, LGNN showed the weakest performance in this scenario, which doesn't involve significant energy transformations.
    
    \item To enhance our understanding of the dataset's validity and the model's effectiveness, we adjusted the speed parameters in the ParticleCar scenario, dividing the tests into low, medium, and high-speed intervals. The outcomes, as presented in Table~\ref{tab:car_compare}, clearly indicate a decline in predictive accuracy with increasing speed. This trend underscores significant limitations in the current models, highlighting a pronounced inability to achieve precise prediction across varying speeds, thereby challenging their efficacy in dynamic scenarios.
\begin{table*}[t]
\renewcommand\arraystretch{1.5}
\caption{Rollout MSE on ParticleCar. L(low-speed), M(medium-speed), and H(high-speed) represent three speed intervals: 10-30, 30-70, and 70-100 }
\centering
\scalebox{0.69}{
\label{tab:car_compare}
\begin{tabular}{l|c|c|c|c|c|c|c|c|c}
\bottomrule
\multicolumn{1}{c|}{\textbf{Model}} & \multicolumn{3}{c|}{\textbf{t = 30}} & \multicolumn{3}{c|}{\textbf{t = 60}} & \multicolumn{3}{c}{\textbf{t = 90}}\\
\cline{2-10}
& \textbf{\textit{L}} & \textbf{\textit{M}} & \textbf{\textit{H}} & \textbf{\textit{L}} & \textbf{\textit{M}} & \textbf{\textit{H}} & \textbf{\textit{L}} & \textbf{\textit{M}} & \textbf{\textit{H}} \\
\hline
\textbf{GNS} & \textbf{1.12$\pm$0.45} & 3.67$\pm$0.85 & 5.08$\pm$1.67 & \textbf{6.79/1.13} & 12.88$\pm$3.71 & 20.32$\pm$5.19 & \textbf{8.97$\pm$1.01} & 34.33$\pm$4.92 & 61.39$\pm$5.09 \\
\textbf{SGNN} & \textbf{1.19$\pm$0.79} & 5.21$\pm$1.47 & 8.01$\pm$2.59 & \textbf{4.51$\pm$1.52} & 15.29$\pm$3.45 & 23.53$\pm$5.04 & \textbf{8.81$\pm$2.35} & 32.61$\pm$2.27 & 57.16$\pm$7.02 \\
\textbf{HGNN} & \textbf{13.41$\pm$2.66} & 15.59$\pm$3.54 & 20.23$\pm$1.67 & \textbf{34.78$\pm$5.13} & 45.01$\pm$7.21 & 50.87$\pm$6.48 & \textbf{71.38$\pm$8.01} & 113.03$\pm$9.79 & 217.65$\pm$13.39 \\
\textbf{LGNN} & \textbf{3.42$\pm$0.45} & 4.31$\pm$2.32 & 7.43$\pm$1.99 & \textbf{8.54$\pm$2.70} & 15.30$\pm$6.15 & 25.11$\pm$6.04 & \textbf{21.97$\pm$4.08} & 75.29$\pm$10.84 & 97.26$\pm$5.75 \\
\bottomrule
\end{tabular}
}
    \vskip -0.1in
\end{table*}

\end{enumerate}
   Through extensive experimentation and scientific analysis conducted by our team, we deduce that many GNS models focus narrowly on single-scenario applications, which severely compromises their generalizability across diverse situations. This reveals a critical gap in the field — the lack of robust methodologies to effectively counter this issue. Additionally, despite significant advancements in predicting physical trajectories, current models face an ongoing challenge: as predictions extend over time, the predictive error tends to not only increase but often does so exponentially, especially in our complex MBDS datasets. 
  In conclusion, the above experiments and analysis indicate that our dataset serves effectively as a benchmark for evaluating the performance of various models. Furthermore, Our dataset reflects the current challenges faced in multi-body dynamics and demonstrates that these models are insufficient to assess the complexity of the dataset, thereby providing opportunities for future work.
\section{Conclusion}

In this paper, we introduce a pioneering Multi-Body Simulation dataset, MBDS, which represents the first of its kind in capturing dynamics involving interactions among multiple bodies or systems, as far as we are aware. This dataset represents a significant improvement over previous datasets, as it is directly generated through adherence to rigorous physical laws by Mujoco\cite{mujoco} and the inclusion of real-world disturbances, rather than relying on computer vision techniques and sensors. MBDS includes a range of common scenarios, and its utility has been thoroughly validated through the implementation of various GNS models. We have also established comprehensive baselines, laying the groundwork for future research in this area. Our future objectives include expanding MBDS both in scale and in the diversity of categories to further propel GNS research. We are confident that MBDS will serve as a valuable and challenging asset for the GNS domain.

\bibliographystyle{unsrt}
\bibliography{ICIC/reference}

\end{document}